\documentclass[journal,twocolumn]{IEEEtran}
\usepackage{slashbox,graphicx,times,amsmath,amssymb,cite,subfigure,multirow,booktabs}
\usepackage[lined,ruled,commentsnumbered]{algorithm2e}
\usepackage[flushleft]{threeparttable}

\DeclareMathOperator*{\argmin}{arg\,min}
 \allowdisplaybreaks

\makeatletter

\newcommand{\Rmnum}[1]{\expandafter\@slowromancap\romannumeral #1@}
\makeatother

\begin{document}
\title{Multi-Tasking Genetic Algorithm (MTGA)\\ for Fuzzy System Optimization}

\author{Dongrui~Wu and         Xianfeng~Tan
\thanks{D.~Wu and X.~Tan are with the Key Laboratory of Image Processing and Intelligent Control (Huazhong University of Science and Technology), Ministry of Education. They are also with the School of Artificial Intelligence and Automation, Huazhong University of Science and Technology, Wuhan, China. Email: drwu@hust.edu.cn, xianfeng\_tan@hust.edu.cn.}
\thanks{Dongrui Wu is the corresponding author.}}

\maketitle

\begin{abstract}
Multi-task learning uses auxiliary data or knowledge from relevant tasks to facilitate the learning in a new task. Multi-task optimization applies multi-task learning to optimization to study how to effectively and efficiently tackle multiple optimization problems simultaneously. Evolutionary multi-tasking, or multi-factorial optimization, is an emerging subfield of multi-task optimization, which integrates evolutionary computation and multi-task learning. This paper proposes a novel and easy-to-implement multi-tasking genetic algorithm (MTGA), which copes well with significantly different optimization tasks by estimating and using the bias among them. Comparative studies with eight state-of-the-art single- and multi-task approaches in the literature on nine benchmarks demonstrated that on average the MTGA outperformed all of them, and had lower computational cost than six of them. Based on the MTGA, a \emph{simultaneous optimization strategy} for fuzzy system design is also proposed. Experiments on simultaneous optimization of type-1 and interval type-2 fuzzy logic controllers for couple-tank water level control demonstrated that the MTGA can find better fuzzy logic controllers than other approaches.
\end{abstract}


\IEEEpeerreviewmaketitle

\section{Introduction}

\IEEEPARstart{M}{ulti}-task learning \cite{Caruana1997,drwuMTALR2019} is a subfield of machine learning, particularly transfer learning \cite{Pan2010,drwuTHMS2017,drwuTFS2017,drwuTNSRE2016}, which uses auxiliary data or knowledge from related/similar tasks to facilitate the learning in a new task. As a result, a learning model for the new task can be built with much less task-specific training data. Or, in other words, with the same amount of task-specific data, a much better model could be trained. In multi-task learning, multiple related learning tasks are performed simultaneously using a (partially) shared model representation. As a result, the common information contained in these related tasks can be exploited to improve the learning efficiency and generalization performance of each task-specific model.

Multi-task optimization (MTO) \cite{Swersky2013,Gupta2016,Ong2016,Da2017} applies multi-task learning to optimization to study how to effectively and efficiently tackle multiple optimization problems simultaneously. Evolutionary multi-tasking \cite{Ong2016}, or multi-factorial optimization (MFO) \cite{Gupta2016}, is an emerging subfield of MTO, which integrates evolutionary computation and multi-task learning. It assumes that each constitutive task has some (positive) influence on the evolutionary process of a single population of individuals, and hence evolving multiple populations from different tasks together simultaneously could be more efficient than evolving each individual task separately. A multi-factorial evolutionary algorithm (MFEA) has recently been proposed in \cite{Gupta2016} and demonstrated promising performance in synthetic and real-world MTO problems. More details about MFEA can be found in the Supplementary Material.

In this paper, we consider single-objective multi-tasking optimization, where every point in the search space maps to a scalar objective value. There have been multiple such approaches in the literature, which mainly focus on the following five aspects:
\begin{enumerate}
\item \emph{How to effectively transfer relevant information between tasks}. Yuan \emph{et al.} \cite{Yuan2016} introduced two new improvements, a new unified representation and a new survivor selection procedure, to the MFEA and demonstrated their effectiveness. Bali \emph{et al.} \cite{Bali2017} proposed a linearized domain adaptation strategy to improve the MFEA. It transforms the search space of a simple task to the search space similar to its constitutive complex task. This high order representative space resembles high correlation with its constitutive task and provides a platform for efficient knowledge transfer via crossover. The proposed LDA-MFEA demonstrated competitive performances against the MFEA. Liew and Ting \cite{Liaw2017} proposes a general framework, the evolution of biocoenosis through symbiosis, for evolutionary algorithms to deal with many-tasking problems, and showed that its performance may be better than the MFEA. Ding \emph{et al.} \cite{Ding2018} proposed decision variable translation and shuffling strategies to facilitate knowledge transfer between optimization problems having different locations of the optimums and different numbers of decision variables, and verified their effectiveness in multi-tasking optimization. Feng \emph{et al.} \cite{Feng2018} used autoencoding to explicitly transfer knowledge across tasks in evolutionary multi-tasking, and demonstrated its performance in both single- and multi-objective multi-task optimization problems. Hashimoto \emph{et al.} \cite{Hashimoto2018} pointed out that the MFEA can be viewed as a special island model, and proposed a simple implementation of evolutionary multi-tasking using the standard island model, which achieved promising performance.
\item \emph{How to allocate computing resources to different tasks}. Gong \emph{et al.} \cite{Gong2019} proposed an evolutionary multitasking algorithm using a dynamic resource allocation strategy, which can dynamically allocate more resources to the more difficult tasks.
\item \emph{How to dynamically adjust the amount of information passed between the tasks}. Wen and Ting \cite{Wen2017} proposed two improvements to the MFEA (parting ways detection and resource reallocation), and showed that they can often result in better solutions, especially when the tasks share low similarity of landscapes.
\item \emph{How to combine with other optimization algorithms}. Cheng \emph{et al.} \cite{Cheng2017} developed a particle swarm optimization based co-evolutionary multi-tasking approach for concurrent global optimization, and demonstrated its performance on synthetic functions and in real-world complex engineering design. Chen \emph{et al.} \cite{Chen2017} proposed an evolutionary multi-tasking single-objective optimization approach based on the cooperative co-evolutionary memetic algorithm. Local search based on the quasi-Newton approach was used to accelerate its convergence.
\item \emph{How to apply to new applications}. Sagarna and Ong \cite{Sagarna2016} applied multi-task evolutionary computation to concurrently searching branches in software tests generation. Tang, Gong and Zhang \cite{Tang2017} used evolutionary multi-tasking to evolve the modular topologies of extreme learning machine classifiers. Li \emph{et al.} \cite{Li2018b} employed evolutionary multi-tasking to optimize multiple sparse reconstruction tasks simultaneously.
\end{enumerate}

To our knowledge, except \cite{Yuan2016}, all other approaches adopt the \emph{unified representation} used in the MFEA, which transforms the solutions of different tasks into a common \emph{unified search space}, and then performs information transfer in this space. A careful examination of the principle of unified representation reveals that the location of the optimum in the unified search space is related to the ranges of the variables. That is, if two tasks have different variable ranges, even if their fitness landscapes are identical in the original space, their optima will be different in the unified search space. In the absence of prior knowledge, correctly setting the variable ranges is very challenging, but is also critical to the success of existing multitasking evolutionary algorithms. Furthermore, even if we can correctly specify the variable ranges, the locations of the optima of different tasks in the unified search space may also be different. Hence, it is very important to consider the bias between different tasks.

To mitigate the negative effects of bias, we should estimate and use it during chromosome transfer. Based on this motivation, we propose a novel multi-tasking genetic algorithm (MTGA), which on average outperforms eight state-of-the-art approaches on nine multi-tasking benchmarks \cite{Da2017}. It also demonstrates outstanding performance in simultaneous optimization of type-1 (T1) and interval type-2 (IT2) fuzzy logic controllers (FLCs).

The main contributions of this paper are:
\begin{enumerate}
\item We propose a novel MTGA approach for multi-task optimization, which has superior performance and low computational cost.
\item We propose a novel MTGA-based simultaneous optimization strategy for fuzzy system design, and demonstrate its effectiveness in fuzzy logic controller optimization.
\end{enumerate}

The remainder of the paper is organized as follows: Section~\ref{sect:MTGA} proposes the MTGA approach. Section~\ref{sect:experiments} validates the performance of the MTGA on nine benchmarks. Section~\ref{sect:FLCs} proposes an MTGA-based simultaneous optimization strategy for fuzzy system design, and demonstrates its performance on simultaneous optimization of T1 and IT2 FLCs for coupled-tank water level control. Section~\ref{sect:conclusions} draws conclusion.

\section{Multi-Tasking Genetic Algorithm (MTGA)} \label{sect:MTGA}

This section introduces our proposed MTGA, whose pseudocode is given in Algorithm~1.

We use the same problem setting as the one in \cite{Gupta2016}. Without loss of generality, assume there are $M$ tasks, all of which are minimization problems. The $m$th task, $T_m$, has an objective function $f_m\, :\, \mathbf{X}_m \rightarrow \mathbb{R}$ on a search space $\mathbf{X}_m$. Each task may also be constrained by several equalities and/or inequalities that must be satisfied by a feasible solution. The goal of the MTGA is to find:
\begin{align}
\{\mathbf{x}_1^*, ..., \mathbf{x}_M^*\} = \{\argmin_{\mathbf{x}_1}f_1(\mathbf{x}_1), ... , \argmin_{\mathbf{x}_M}f_M(\mathbf{x}_M)\},
\end{align}
where $\mathbf{x}_m$ is a feasible solution in $\mathbf{X}_m$, $m=1,...,M$.

For the ease of illustration, we only consider two tasks, i.e., $M=2$. The extension to more than two tasks is straightforward.

\begin{algorithm}[!h] 
\KwIn{Two tasks $T_1$ and $T_2$\;
\hspace*{10mm} $N$, the population size\;
\hspace*{10mm} $K$, the maximum number of generations\;
\hspace*{10mm} $n_t$, the number of transferred chromosomes.}
\KwOut{$\mathbf{x}_m^*$, the best chromosome for each $T_m$, $m=1,2$.}
\For{$m=1,2$}{
Randomly generate $N$ chromosomes $\{\mathbf{x}_{m,n}\}_{n=1}^N$ to initialize the population $P_m$ for Task $T_m$\;
Compute the fitness for each $\mathbf{x}_{m,n}$ in $P_m$\;
Sort all chromosomes in $P_m$ in descending order according to their fitness\;
}
Set $k=1$\;
\While{$k\le K$}{
Compute $\mathbf{m}_1$ and $\mathbf{m}_2$ in (\ref{eq:m})\;
\For{$m=1,2$}{
Initialize a temporary population $P_t=\emptyset$\;
Construct the first $n_t$ chromosomes of $P_t$ from the $n_t$ best chromosomes in $P_{3-m}$, using (\ref{eq:bias})\;
Construct the remaining $N-n_t$ chromosomes of $P_t$ as the $N-n_t$ best chromosomes in $P_m$\;
Construct an index vector $\mathbf{s}$ as a random permutation of $[1,...,N]$\;
Initialize the offspring population $O=\emptyset$\;
\For{$n=1,...,N/2$}{
Pick two parents $\mathbf{x}_{m,\mathbf{s}(n)}$ and $\mathbf{x}_{m,\mathbf{s}(N/2+n)}$\;
Crossover to generate two offsprings $\mathbf{x}_e$ and $\mathbf{x}_f$ according to (\ref{eq:xe}) and (\ref{eq:xf})\;
Mutate each of $\mathbf{x}_e$ and $\mathbf{x}_f$\;
Add $\mathbf{x}_e$ and $\mathbf{x}_f$ to $O$\;}
Evaluate the fitness of each chromosome in $O$\;
Set $P=P_m\cup O$\;
Sort the chromosomes in $P$ in descending order according to their fitness\;
Form the new population $P_m$ using the $N$ fittest chromosomes in $P$\;}
$k=k+1$\;}
\textbf{Return} The fittest chromosome for each $T_m$, $m=1,2$.
\caption{Pseudocode of the proposed MTGA.}
\end{algorithm}

\subsection{Motivation of the Proposed MTGA}

The MFEA may offer no advantage over optimizing each single task separately, if in the unified search space the optimal solutions of different tasks, or their fitness landscapes, are significantly different. Unfortunately, in practice often we do not know \emph{a priori} how similar the tasks are, and it is desirable to have a multi-tasking optimization algorithm that can achieve good performance even in the worst-case scenario that the tasks are significantly different.

The MTGA is proposed to cope with the above problem. Particularly, it addresses two important questions: 1) how to estimate the difference between the optimal solutions of the two tasks, and, 2) how to effectively transfer the fittest chromosomes between the two populations (tasks). Its main idea is to estimate the bias between the two tasks and then remove it in chromosome transfer, so that the optimal solutions of the two tasks are close to each other. In this way, a promising chromosome from one task can also be transformed into a promising solution for the other task, expediting the convergence.

\begin{figure}[h]\centering
\includegraphics[width=.9\linewidth,clip]{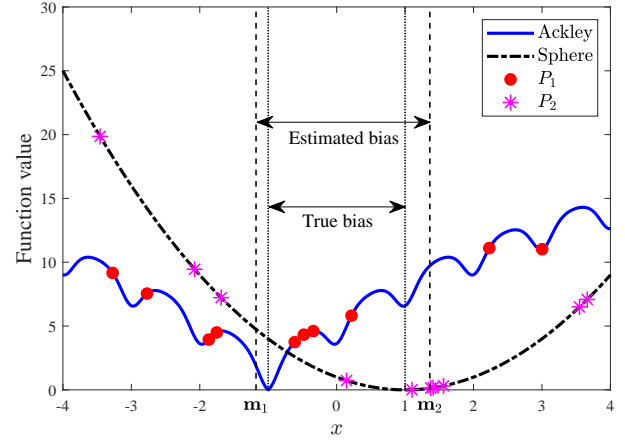}
\caption{Illustration of how the MTGA estimates the bias between two tasks. $P_1$ ($P_2$) is the population of 1-D Ackley (Sphere). $\mathbf{m}_1$ ($\mathbf{m}_2$) is the mean of the fittest four chromosomes in $P_1$ ($P_2$). $|\mathbf{m}_2 - \mathbf{m}_1|$ is the estimated bias.} \label{fig:bias}
\end{figure}

The MTGA's approach to estimate the bias can be explained using the example in Fig.~\ref{fig:bias}, where the two tasks are one-dimensional Ackley and Sphere functions \cite{Da2017}. Ackley maintains a population $P_1$, and Sphere a population $P_2$, each of which has 10 chromosomes. Clearly, there is a large bias between their optimal solutions, and hence blindly transferring a promising solution from $P_1$ to $P_2$ (or the opposite) may not benefit the search. The MTGA first computes the mean of a few fittest chromosomes (four were used in our example) in each population, and then estimates the bias between their optimal solutions as the difference between these two means. When transferring a promising chromosome from one population to the other, it adds (or subtracts, depending on the direction) this bias to make it more compatible with the new task. In the first few generations, bias estimation may not be very accurate, because the fittest chromosomes in each population may be far away from its global optimum. However, as the evolution goes on, the fittest chromosomes will move towards their corresponding global optima, and hence the bias estimate will be more accurate, which benefits the transfer more.

\begin{figure}[h]\centering
\includegraphics[width=.9\linewidth,clip]{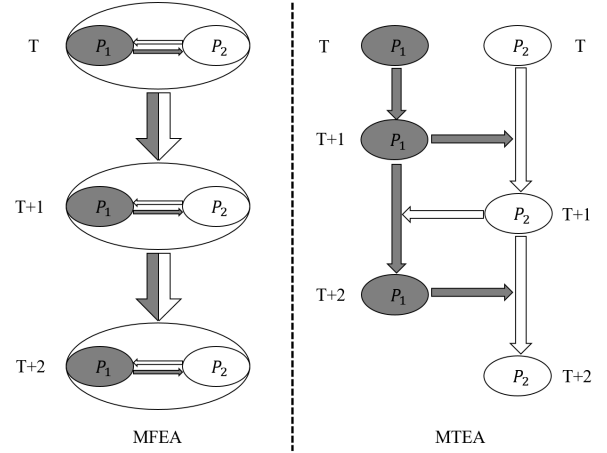}
\caption{Illustration of the workflow difference between the MFEA and the MTGA. $T$ is the index of the evolutionary generation.} \label{fig:MTGA}
\end{figure}

To effectively transfer the fittest chromosomes between the two populations, the MTGA uses sequential transfer in each generation, i.e., $P_1$ first transfers its fittest chromosomes to $P_2$ (after considering the bias), then $P_2$ performs crossover, mutation, fitness evaluation and reproduction to generate a new $P_2$. The fittest chromosomes in this new $P_2$ are then transferred to $P_1$ (after considering the bias), which next goes through crossover, mutation, fitness evaluation and reproduction to generate a new $P_1$. In this way, the updated fittest chromosomes in one population are immediately used by the other in the same generation, expediting the converge. On the contrary, the MFEA uses simultaneous chromosome transfer, i.e., the two tasks transfer their fittest chromosomes to the other simultaneously (without considering the bias), and then perform crossover, mutation, fitness evaluation and reproduction separately. As a result, the updated fittest chromosomes for one task cannot be shared by the other until the next generation, which is a waste of information. The difference is illustrated in Fig.~\ref{fig:MTGA}.

The details of the MTGA are presented next.

\subsection{Population Initialization}

Unlike the MFEA, which keeps a single population and uses a skill factor to identity which chromosome belongs to which task, the MTGA keeps a separate population for each individual task.

Let $N$ be the population size of each task, and $P_m$ the population for $T_m$ ($m=1,2$). $P_m$ is randomly initialized in the first generation of the MTGA.

\subsection{Chromosome Transfer}

Chromosome transfer is considered in each subsequent iteration. We first compute the means of the best $n_t$ chromosomes in $P_1$ and $P_2$ respectively, and denote them as $\mathbf{m}_1$ and $\mathbf{m}_2$. The difference between $\mathbf{m}_1$ and $\mathbf{m}_2$ represents the bias between the two tasks, which can be used to make the transferred chromosomes more consistent with the new population.

Let's focus on $T_1$ as an example. We transfer $n_t$ best chromosomes from $P_2$ to $P_1$, to replace the $n_t$ worst chromosomes in $P_1$. Assume the chromosomes in $P_1$ and $P_2$ have been sorted from the best to the worst according to their fitness, respectively. Denote the sorted chromosomes as $\{\mathbf{x}_{1,n}\}_{n=1}^N$ and $\{\mathbf{x}_{2,n}\}_{n=1}^N$, respectively. Then,
\begin{align}
\mathbf{m}_1=\frac{1}{n_t}\sum_{n=1}^{n_t}\mathbf{x}_{1,n}, \quad
\mathbf{m}_2=\frac{1}{n_t}\sum_{n=1}^{n_t}\mathbf{x}_{2,n} \label{eq:m}
\end{align}
We then construct a temporary population $P_t$ as:
\begin{align}
P_t=\{\mathbf{x}_{1,1}', ..., \mathbf{x}_{1,n_t}', \mathbf{x}_{1,1},...,\mathbf{x}_{1,N-n_t}\},
\end{align}
where $\mathbf{x}_{1,n}'$ ($n=1,...,n_t$) is a transferred chromosome from $P_2$, computed as follows.

Let $d_m=|\mathbf{X}_m|$ be the dimensionality of the search space of Task $T_m$, or equivalently, the number of genes in a chromosome in $P_m$, $m=1,2$. When $d_1\ge d_2$, we construct an index set $I$ by randomly sampling \emph{without} replacement $d_2$ locations from the $d_1$ locations. When $d_1< d_2$, we construct an index set $I$ by randomly sampling \emph{with} replacement $d_2$ locations from the $d_1$ locations. Then,
\begin{align}
\mathbf{x}_{1,n}'(i)=\mathbf{x}_{2,n}(I(i))-\mathbf{m}_2(I(i))+\mathbf{m}_1(i), \label{eq:bias}
\end{align}
where $\mathbf{x}_{1,n}'(i)$ is the $i$th gene in $\mathbf{x}_{1,n}'$, $\mathbf{x}_{2,n}(I(i))$ is the $I(i)$th gene in $\mathbf{x}_{2,n}$, $\mathbf{m}_2(I(i))$ is the $I(i)$th gene in $\mathbf{m}_2$, $\mathbf{m}_1(i)$ is the $i$th gene in $\mathbf{m}_1$, $n=1,...,n_t$, and $i=1,...,d_1$.

Note that $I$ is usually randomly constructed for each transferred chromosome, i.e., the $i$th gene in $\mathbf{x}_{1,n}'$ is randomly matched to a gene (not necessarily the $i$th gene) in $\mathbf{x}_{2,n}$, and the matching is also different for different $n$. We use a random matching instead of a fixed matching because usually in practice we do not know which gene in $T_2$ can best benefit a gene in $T_1$, and randomizing the matching for different genes and different $n$ offers more diversity and higher likelihood to find a good matching than a blind fixed matching. However, if we know there is a correspondence between a pair of genes, then a fixed matching may also be used.

Once a matching between $\mathbf{x}_{1,n}'(i)$ and $\mathbf{x}_{2,n}(I(i))$ is established, $\mathbf{m}_1(i)-\mathbf{m}_2(I(i))$ represents the bias between the two genes in the two tasks, and subtracting this bias from $\mathbf{x}_{2,n}(I(i))$ makes the genes in the two tasks more consistent, and hence facilitates the knowledge transfer.

Note also that there could be optimization problems in which some gene values are essentially categorical, and hence (\ref{eq:m}) should not be applied to them. For example, for the fuzzy system optimization problem introduced in the next section, we may also want to optimize the rulebase so that good control performance may be obtained by fewer than nine rules, instead of always using the nine rules in Table~\ref{tab:rules}. One approach is to add nine new genes to the chromosome, each indicating whether a particular rule in Table~\ref{tab:rules} should be used or not. These genes may take values of 0 (the corresponding rule is not used) or 1 (the corresponding rule is used), but we cannot simply compute their average, e.g., the average of 0 and 1 is 0.5, which is not meaningful. In this case, we should exclude these genes from chromosome transfer, and consider only the genes which are truly numerical.

\subsection{Crossover and Mutation}

Next, we perform crossover, and make sure the $n_t$ transferred chromosomes from $P_2$ are all used in the crossover.

We define an index vector $\mathbf{s}$ as a random permutation of $[1,...,N]$. Each time, we pick two parents $\mathbf{x}_{\mathbf{s}(n)}$ and $\mathbf{x}_{\mathbf{s}(n+N/2)}$ ($n=1,...,N/2$), and use the simulated binary crossover (SBX) \cite{Deb1994} and polynomial mutation \cite{Deb1999} operators, the same as those in \cite{Da2017}.

For notation simplicity, let the two parents be $\mathbf{x}_{\mathbf{s}(n)}=[\mathbf{x}_a(1),...,\mathbf{x}_a(d)]$ and $\mathbf{x}_{\mathbf{s}(n+N/2)}=[\mathbf{x}_b(1),...,\mathbf{x}_b(d)]$, where $d$ is the dimensionality of the search space. Then, in SBX, we first compute:
\begin{align}
\mathbf{c}(j)=\left\{\begin{array}{ll}
               (2r)^{1/(\beta+1)}, & r\le 0.5 \\
               \left[2(1-r)\right]^{-1/(\beta+1)}, & r>0.5
             \end{array}\right.,\quad j=1,...,d \label{eq:SBX}
\end{align}
where $\beta$ is a user-specified parameter, and $r$ is a random number in $[0,1]$, which is regenerated for each $j$. The two offsprings, $\mathbf{x}_e=[\mathbf{x}_e(1),...,\mathbf{x}_e(d)]$ and $\mathbf{x}_f=[\mathbf{x}_f(1),...,\mathbf{x}_f(d)]$, obtained from the SBX are ($j=1,...,d$):
\begin{align}
\mathbf{x}_e(j)=[(1+\mathbf{c}(j))\mathbf{x}_a(j) + (1-\mathbf{c}(j))\mathbf{x}_b(j)]/2, \label{eq:xe}\\
\mathbf{x}_f(j)=[(1+\mathbf{c}(j))\mathbf{x}_b(j) + (1-\mathbf{c}(j))\mathbf{x}_a(j)]/2, \label{eq:xf}
\end{align}
Clearly, $\mathbf{x}_e+\mathbf{x}_f=\mathbf{x}_{\mathbf{s}(n)}+\mathbf{x}_{\mathbf{s}(n+N/2)}$. Additionally, it's easy to observe that $\mathbf{x}_e$ is closer to $\mathbf{x}_{\mathbf{s}(n)}$ than to $\mathbf{x}_{\mathbf{s}(n+N/2)}$, and $\mathbf{x}_f$ is closer to $\mathbf{x}_{\mathbf{s}(n+N/2)}$ than to $\mathbf{x}_{\mathbf{s}(n)}$, because $\mathbf{c}(j)>0$.

Let $\eta$ be a user-specified parameter, and $r$ be a random number in $[0,1]$. Then, the polynomial mutation of a gene $\mathbf{x}(j)$ with range $[l,u]$ is computed as \cite{Deb1999}:
\begin{align}
\mathbf{x}'(j)=\left\{\begin{array}{ll}
              \mathbf{x}(j)+[(2r)^{\frac{1}{1+\eta}}-1](\mathbf{x}(j)-l), & r\le 0.5 \\
              \mathbf{x}(j)+[1-(2(1-r))^{\frac{1}{1+\eta}}](u-\mathbf{x}(j)), & r>0.5
            \end{array}\right. \label{eq:PM}
\end{align}

After mutation, $\mathbf{x}_e$ and $\mathbf{x}_f$ are added to the offspring population $O$. The above crossover and mutation operations are repeated $N/2$ times so that $O$ has $N$ chromosomes.

\subsection{Reproduction}

We then evaluate the fitness of each chromosome in $O$, combine the chromosomes in $O$ with those in $P_m$, and sort the $2N$ chromosomes from the best to the worse according to their fitness. We use an elitist selection mechanism to prorogate the first $N$ best chromosomes to the next generation for $T_m$.

\section{Experiments on Benchmarks} \label{sect:experiments}

This section compares the MTGA with eight state-of-the-art single- and multi-task evolutionary algorithms on the nine benchmarks introduced in \cite{Da2017}.

\subsection{Performance Measures}

In addition to the error (the difference from the known minimum) in each task, the simple performance metric proposed in \cite{Da2017} is also used to quantify the performance of different algorithms.

Assume there are $K$ algorithms, $A_1,...,A_K$ for a problem with $M$ minimization tasks $T_1,...,T_M$, and each algorithm has been  run for $L$ repetitions. Let $B(k,m)_l$ denote the best obtained result on the $l$th repetition by Algorithm $A_k$ on Task $T_m$, and $\mu_m$ and $\sigma_m$ be the mean and standard deviation (std) of  $B(k,m)_l$, $k=1,...,K$, $l=1,...,L$. Then, the normalized performance $B'(k,m)_l$ is computed as:
\begin{align}
B'(k,m)_l=\frac{B(k,m)_l-\mu_m}{\sigma_m}
\end{align}
and the \emph{performance score} of Algorithm $A_k$ is:
\begin{align}
s_k=\sum_{m=1}^M\sum_{l=1}^LB'(k,m)_l
\end{align}
A smaller performance score indicates a better overall performance.

\subsection{Algorithms}

We compare the performance of our proposed MTGA with a classic single-task evolutionary algorithm, and seven state-of-the-art multi-task algorithms:
\begin{enumerate}
\item The single-objective evolutionary algorithm (SOEA), which considers each task independently. We used the SOEA code (in Matlab) provided in the WCCI2018 competition on evolutionary multi-task optimization. Essentially, each SOEA implements a genetic algorithm \cite{Godlberg1989,drwuEAAI2006}, with SBX crossover and polynomial mutation.
\item The MFEA \cite{Gupta2016}. We also used the MFEA code (in Matlab) provided in the WCCI2018 competition on evolutionary multi-task optimization.
\item The evolution of biocoenosis through symbiosis (EBS) algorithm \cite{Liaw2017}, which can deal with many-tasking problems. The basic evolutionary algorithm used in the EBS was identical to the genetic algorithm in the SOEA and the MFEA.
\item The MFEA-LBS \cite{Yuan2016}, which employs a permutation based unified representation and level-based selection (LBS) to enhance the original MFEA.
\item The multi-factorial evolutionary algorithm with resource reallocation (MFEARR) \cite{Wen2017}, which adds a resource allocation mechanism to facilitate the discovery and utilization of synergy among tasks.
\item The linearized domain adaptation multi-factorial evolutionary algorithm (LDA-MFEA) \cite{Bali2017}, which uses LDA to transform the search space of a simple task to a new one similar to its constitutive complex task for efficient problem solving.
\item The generalized multi-factorial evolutionary algorithm (G-MFEA) \cite{Ding2018}, which uses decision variable translation and shuffling strategies to facilitate knowledge transfer between optimization problems.
\item The evolutionary multi-tasking via explicit auto-encoding (EMEA) algorithm \cite{Feng2018}, which uses autoencoding to explicitly transfer knowledge across tasks in evolutionary multi-tasking.
\end{enumerate}
We implemented the last six algorithms in Matlab according to the corresponding publications, and tried our best to optimize them.

We used a population size of 200 in the MFEA and its variants. For the SOEA, the G-MTGA, the EMEA and the MTGA, each task had a population size of 100. The maximum number of function evaluations was 100,000 for all algorithms, i.e., all algorithms terminated after 500 iterations. $rmp=0.3$, $\beta=2$ in (\ref{eq:SBX}) of the SBX, and $\eta=5$ in (\ref{eq:PM}) of polynomial mutation, were used in all algorithms. Additionally, $n_t=10$ was used in the EMEA (as in \cite{Feng2018}), and $n_t=40$ in the MTGA.

To cope with the randomness, each algorithm was run 20 times, each time with a randomly initialized population. Then several statistics, such as the mean and std of the objectives in the two tasks, were computed.

\subsection{Experimental Results}

\begin{figure*}[!h]\centering
\subfigure[]{\label{fig:O1}\includegraphics[width=.49\linewidth,clip]{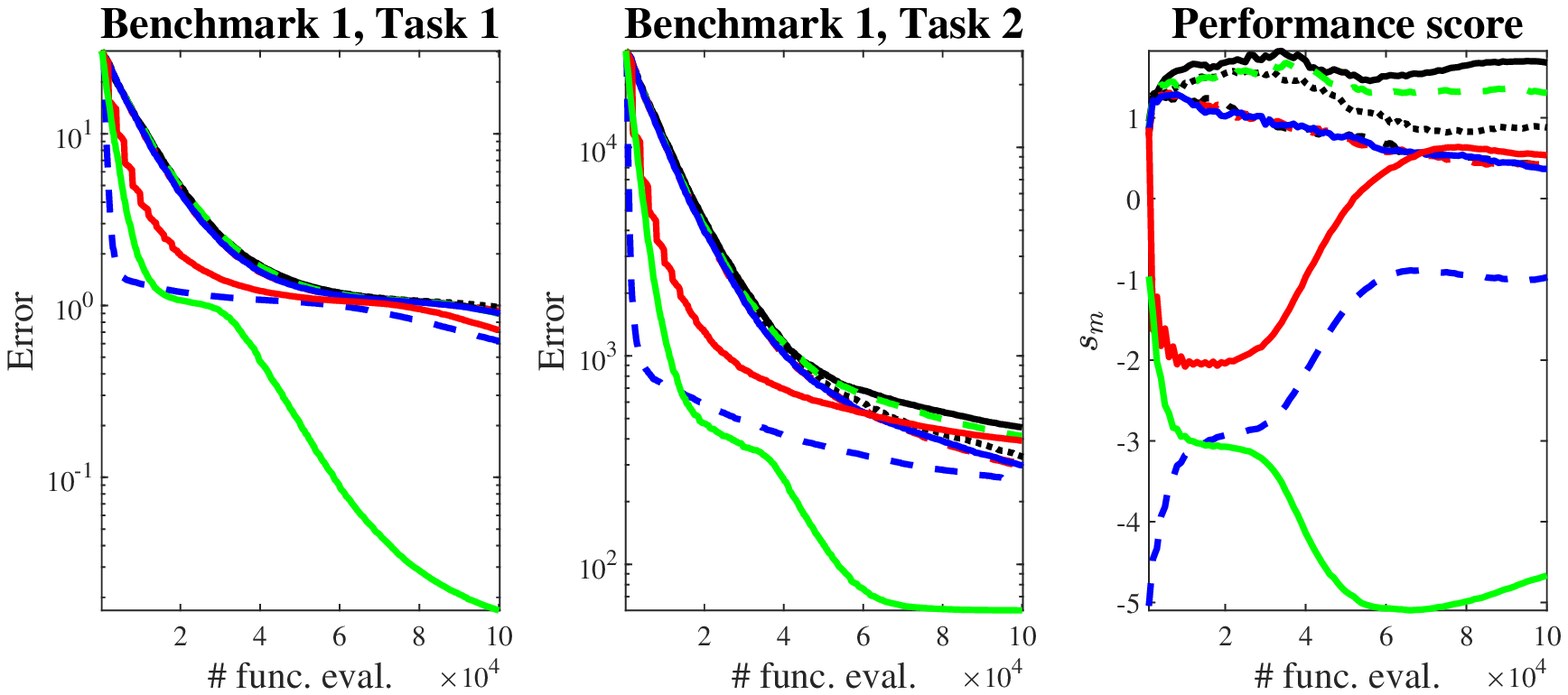}}
\subfigure[]{\label{fig:O2}\includegraphics[width=.49\linewidth,clip]{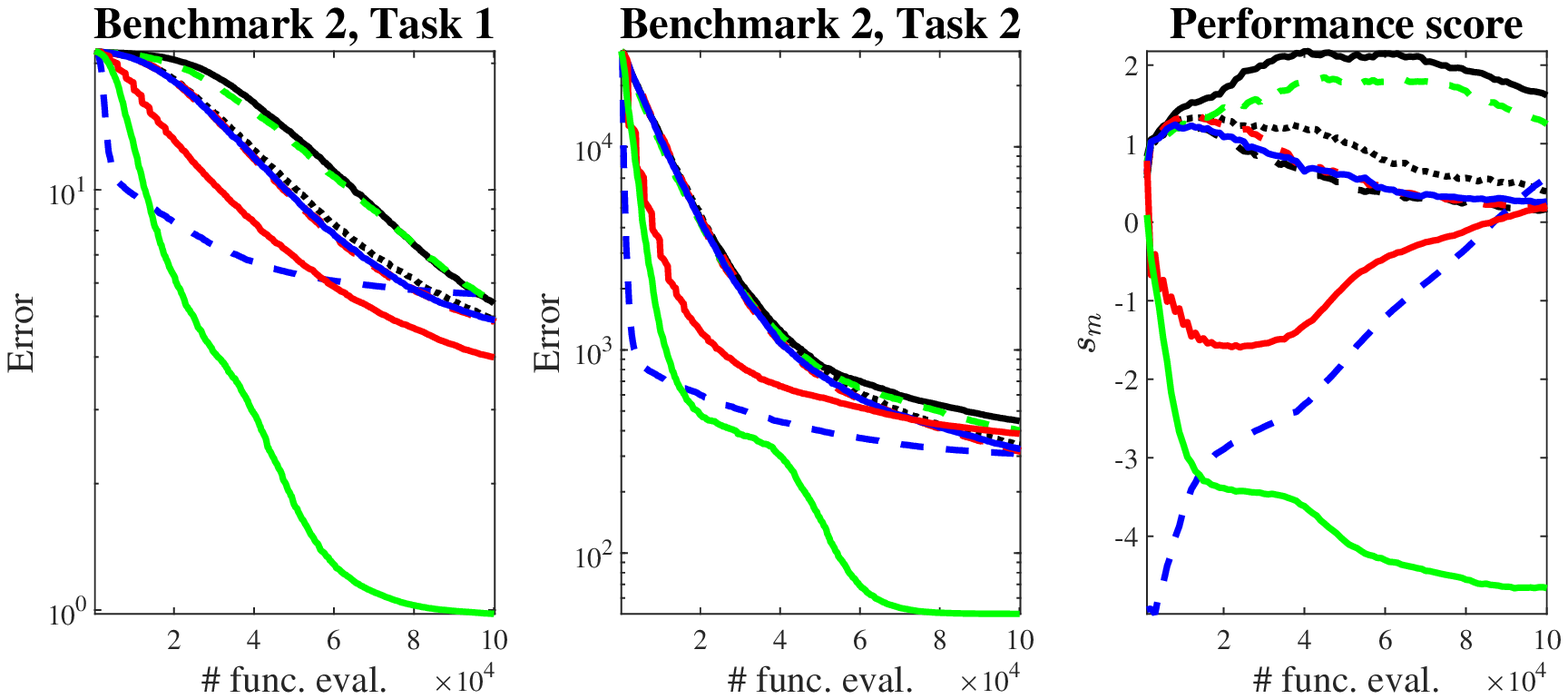}}
\subfigure[]{\label{fig:O3}\includegraphics[width=.49\linewidth,clip]{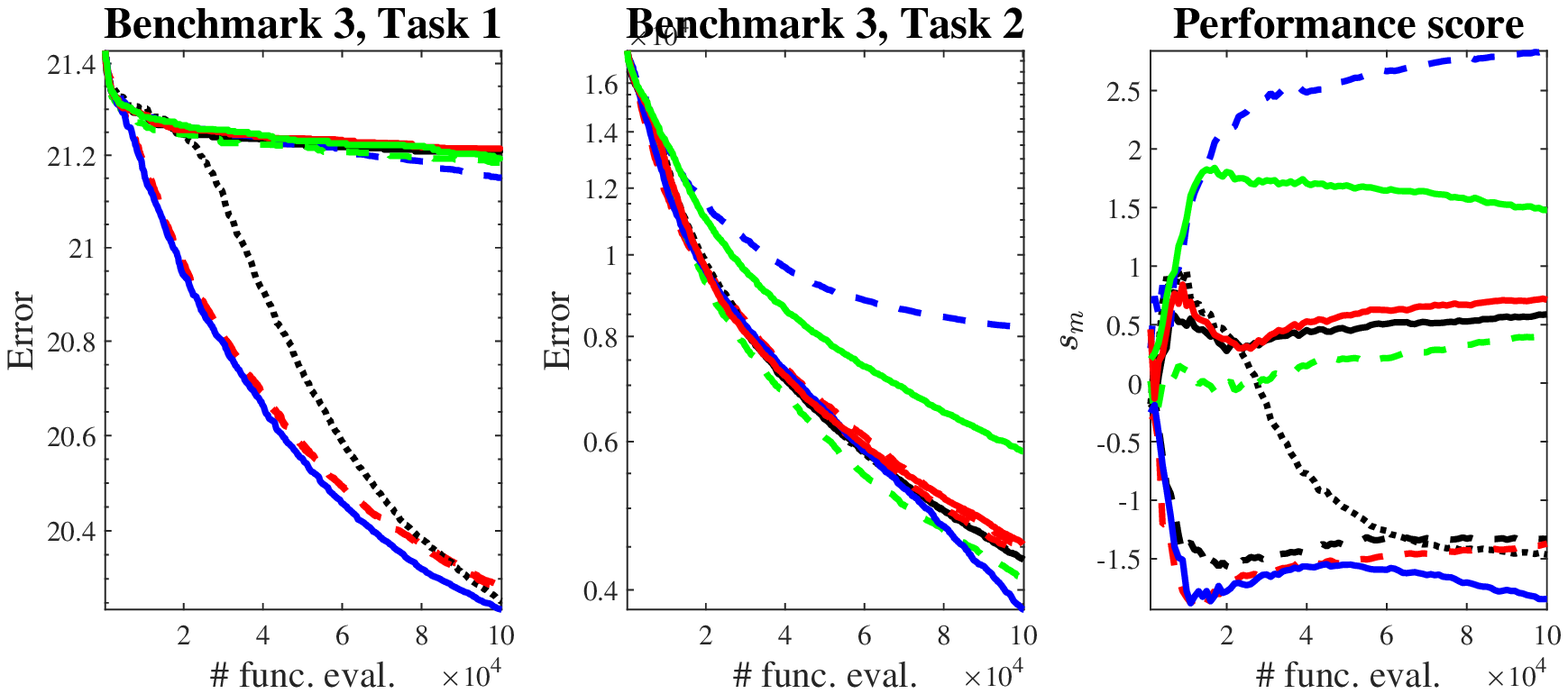}}
\subfigure[]{\label{fig:O4}\includegraphics[width=.49\linewidth,clip]{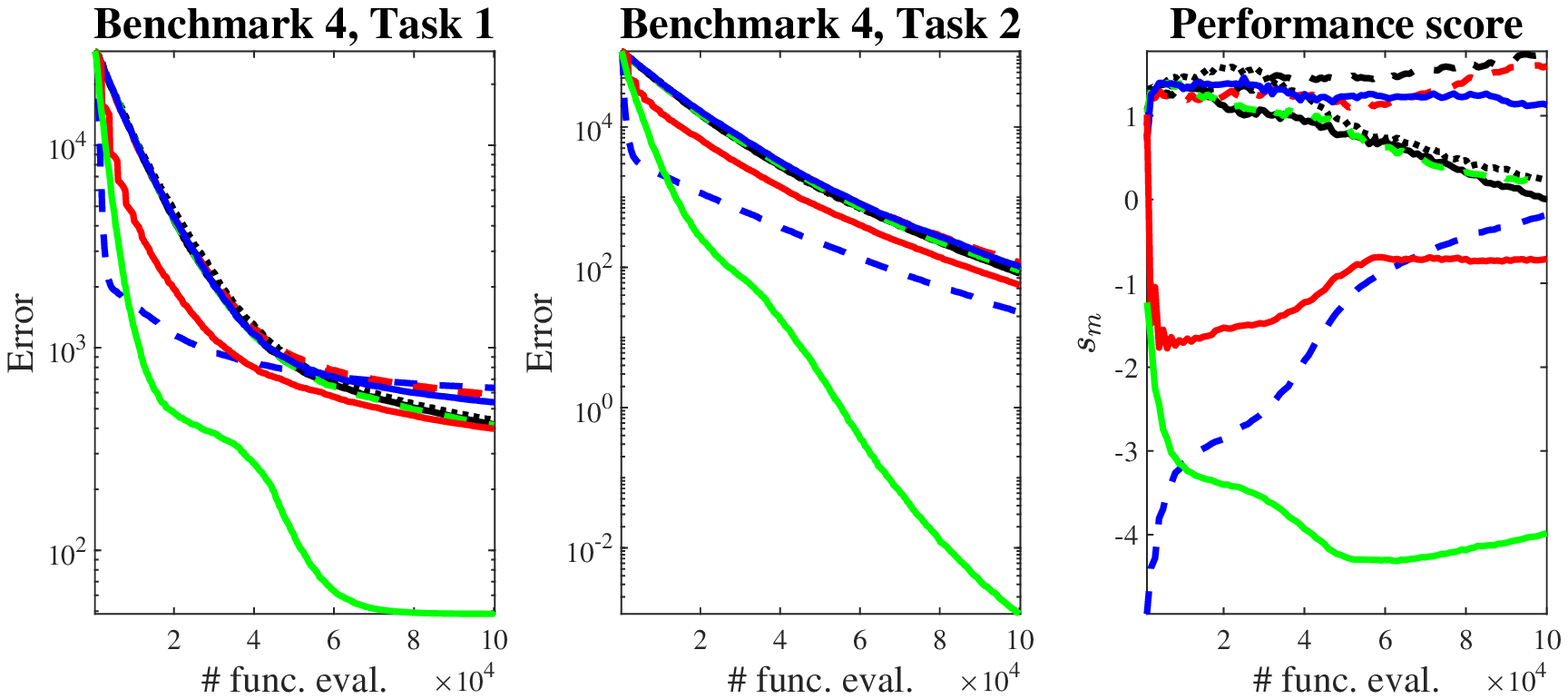}}
\subfigure[]{\label{fig:O5}\includegraphics[width=.49\linewidth,clip]{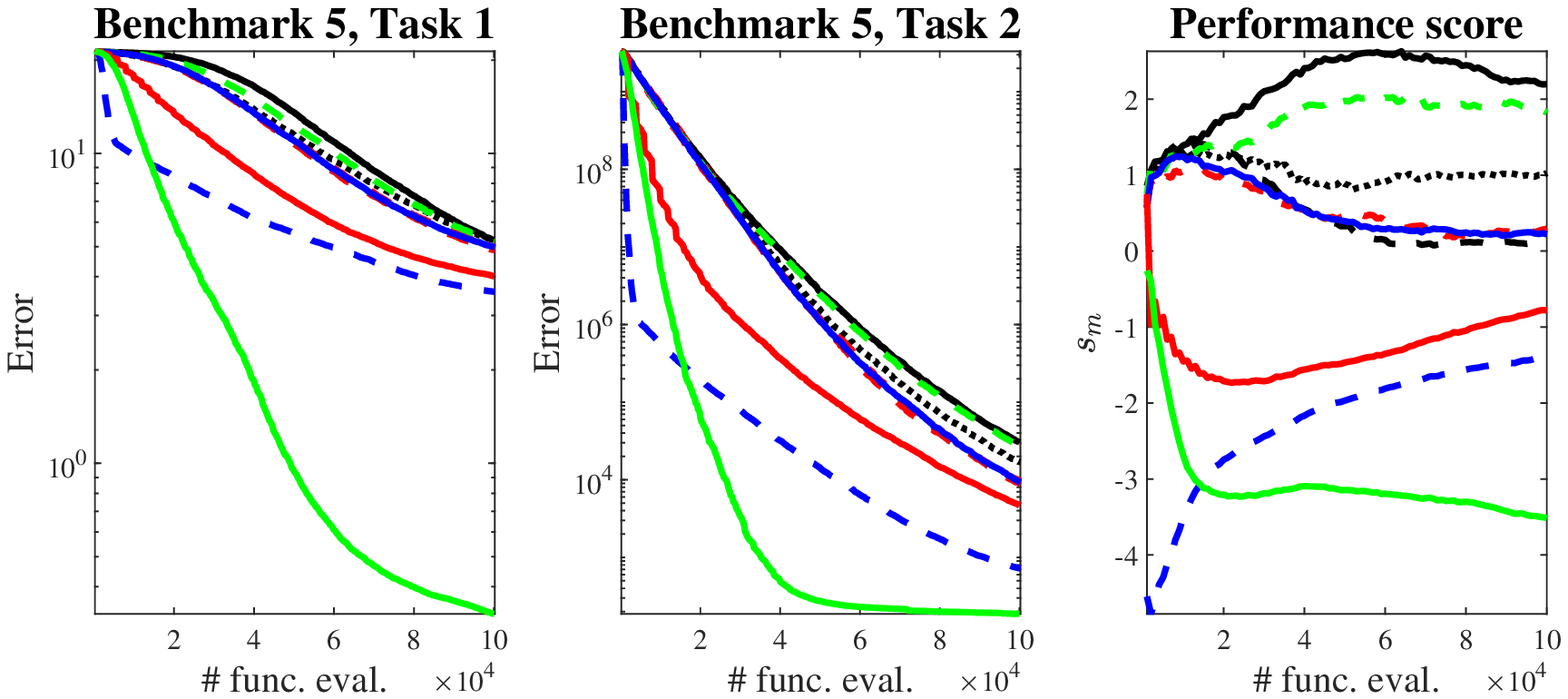}}
\subfigure[]{\label{fig:O6}\includegraphics[width=.49\linewidth,clip]{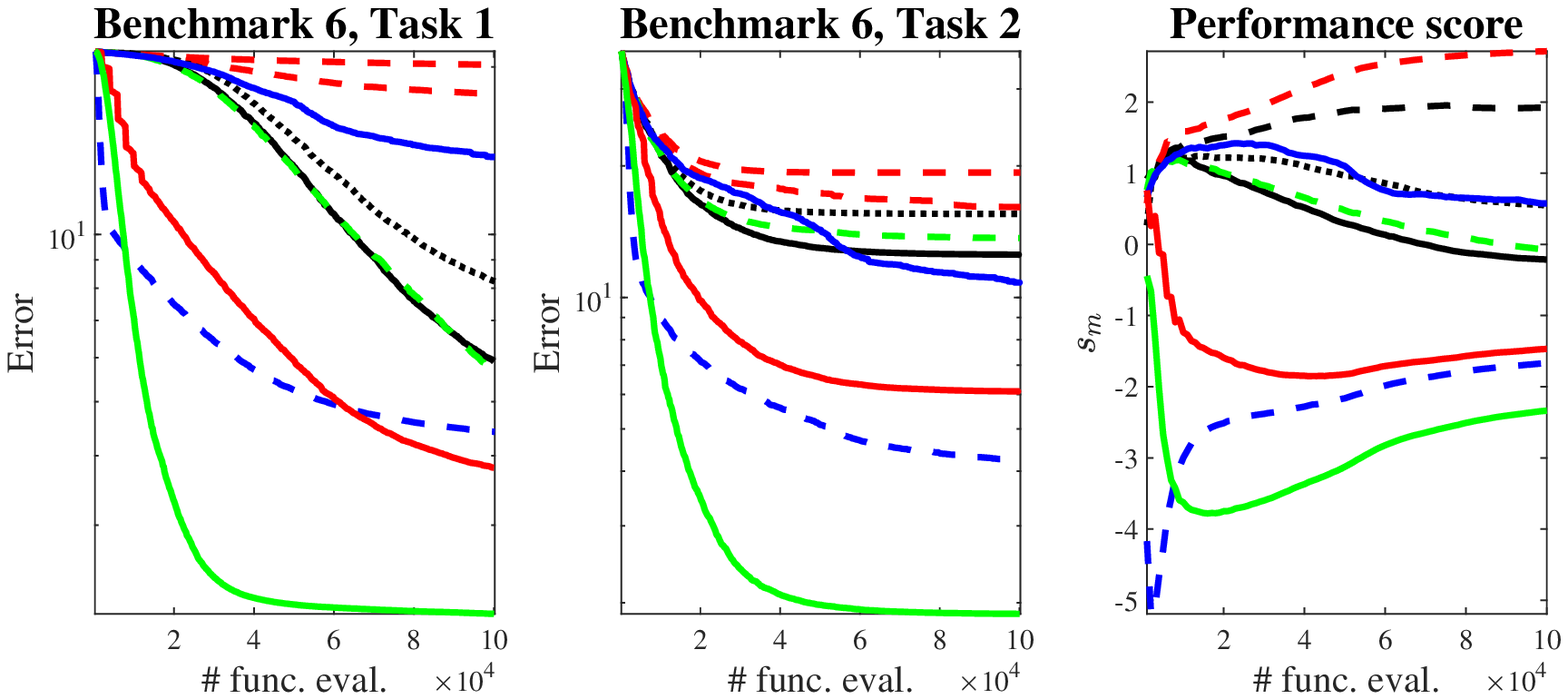}}
\subfigure[]{\label{fig:O7}\includegraphics[width=.49\linewidth,clip]{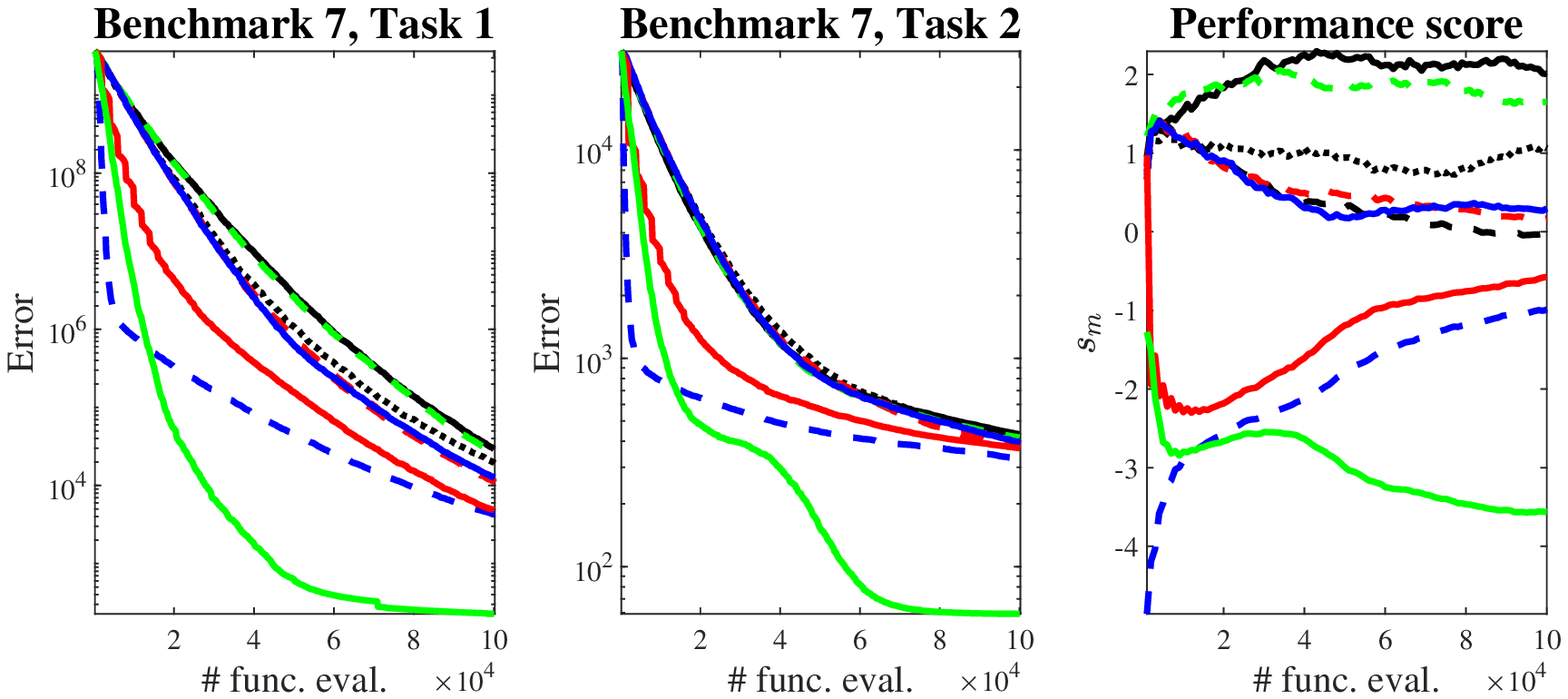}}
\subfigure[]{\label{fig:O8}\includegraphics[width=.49\linewidth,clip]{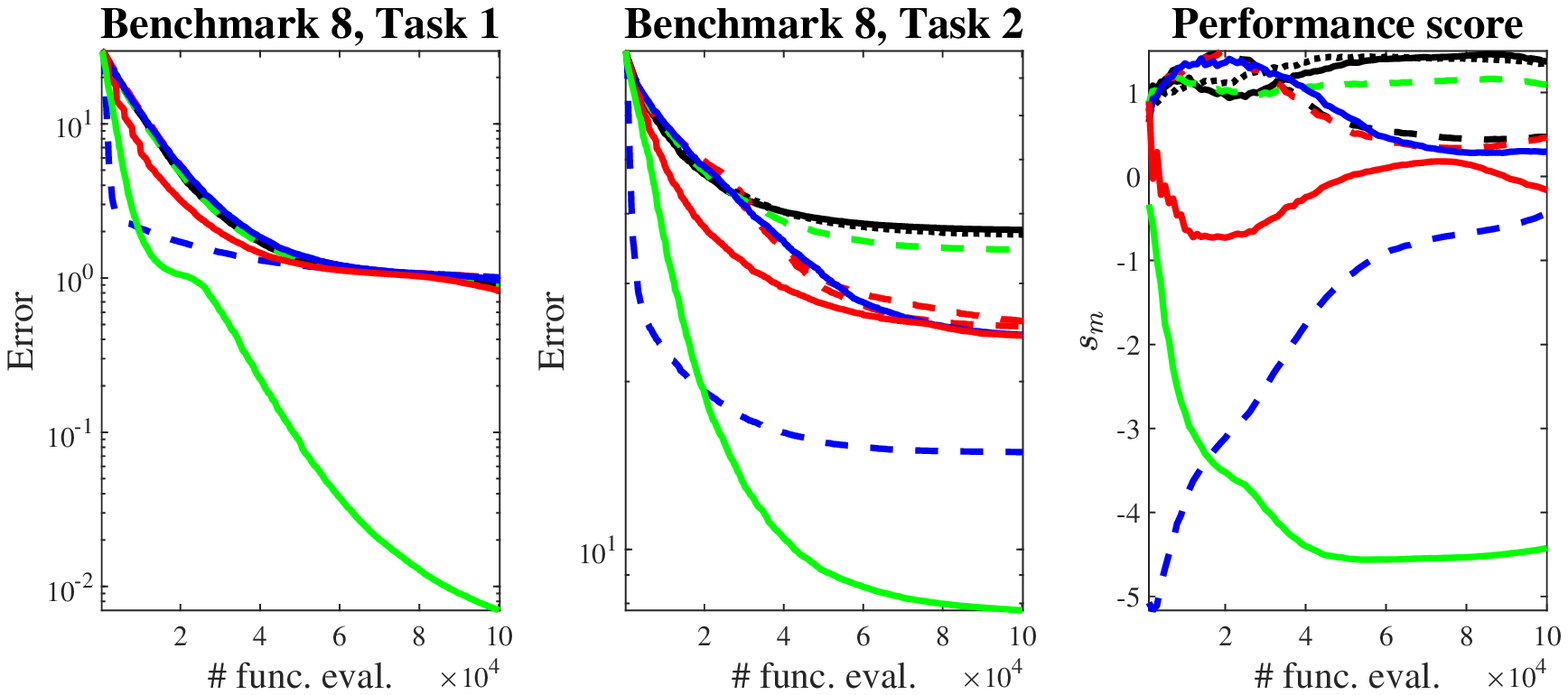}}
\subfigure[]{\label{fig:O9}\includegraphics[width=.58\linewidth,clip]{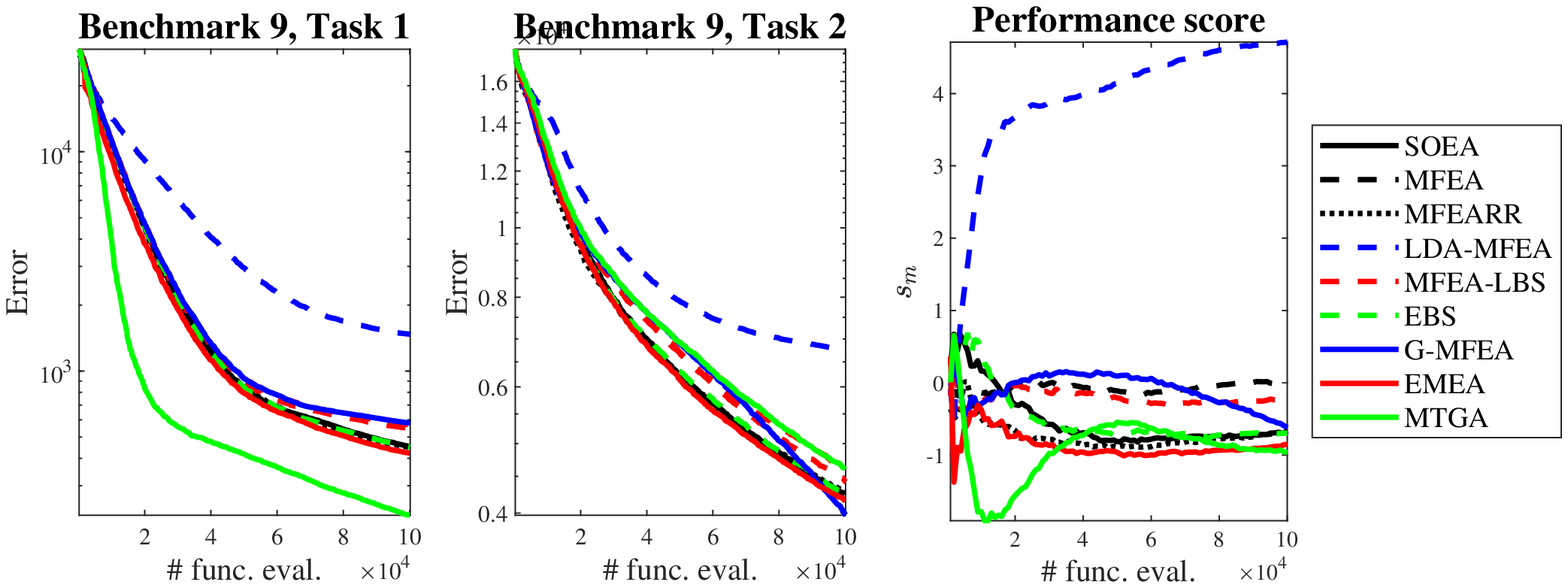}}
\caption{Experimental results on the nine benchmarks.} \label{fig:O}
\end{figure*}

\begin{figure}[h]\centering
\includegraphics[width=\linewidth,clip]{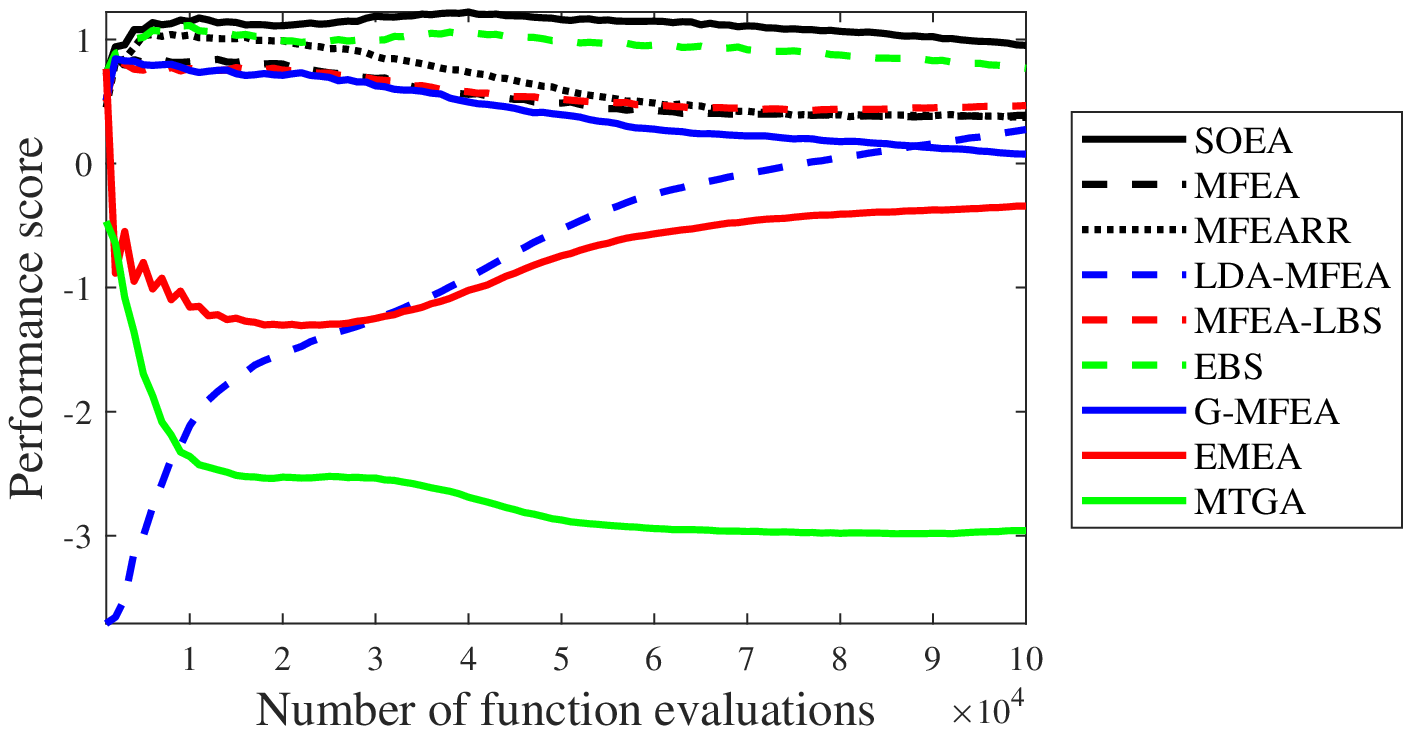}
\caption{Average performance scores across the nine benchmarks.} \label{fig:Oscore}
\end{figure}

Individual experimental results on the nine benchmarks are shown in Fig.~\ref{fig:O}, and the average performance scores of the nine algorithms across the nine benchmarks are shown in Fig.~\ref{fig:Oscore}. Observe that:
\begin{enumerate}
\item On average all multi-task algorithms outperformed the SOEA, which suggests that the transfer of information between the tasks can indeed improve the overall optimization performance.
\item On average the EBS and the MFEARR performed worse than the MFEA, and the MFEA-LBS had comparable performance as the MFEA.
\item The LDA-MFEA achieved the best performance when the number of function evaluations was small, but gradually degraded when the number of function evaluations increased.
\item The G-MFEA slightly outperformed the MFEA when the number of function evaluations was large.
\item Among the eight existing algorithms, on average EMEA achieved the second best performance (worse only than the LDA-MFEA) when the number of function evaluations was small, and the best performance when the number of function evaluations was large.
\item Among all nine algorithms, on average our proposed MTGA achieved the second best performance (worse only than the LDA-MFEA) when the number of function evaluations was small, and the best performance when the number of function evaluations was large.
\end{enumerate}

The average errors (mean and std) of different algorithms on different tasks, after 100,000 function evaluations, are given in Table~\ref{tab:resultsO}. Observe that the MTGA achieved the best performance score in eight out of the nine benchmarks. Overall it dominated the other eight algorithms. More experimental results on the parameter sensitivity of the MTGA, and its performance on nine more challenging and practical benchmarks, can be found in the Supplementary Material.

\begin{table*}[h]
\caption{Average errors (mean and parenthesized std) of different algorithms on the nine benchmarks, after 100,000 function evaluations. Best performances are marked in bold.}
\centering \setlength{\tabcolsep}{1.2mm}
\begin{tabular}{c|ccccccccccc}   \hline
&\multicolumn{2}{c}{Benchmark}& 1 & 2 & 3 & 4 & 5 & 6 & 7 & 8 & 9 \\ \hline
         & \multirow{2}{*}{T1}  &mean      & $0.8996$ & $5.3695$ & $21.2084$ & $417.8177$ & $5.2545$ & $5.9086$ & $29341.5108$ & $0.9121$ & $452.8721$ \\
         &                      &std       & $(0.0685)$ & $(1.0510)$ & $(0.0397)$ & $(50.3642)$ & $(0.6350)$ & $(3.5919)$ & $(13428.4702)$ & $(0.0530)$ & $(63.0397)$ \\ \cline{2-12}
SOEA     & \multirow{2}{*}{T2}  &mean     & $453.9618$ & $446.0697$ & $4350.4544$ & $81.9612$ & $30552.1236$ & $12.5301$ & $431.0700$ & $37.4077$ & $4261.6635$ \\
         &                      &std       & $(54.2633)$ & $(50.7552)$ & $(567.0568)$ & $(23.4086)$ & $(12233.1876)$ & $(2.5087)$ & $(51.2238)$ & $(4.2753)$ & $(438.0050)$ \\ \cline{2-12}
         & Score                &           & $1.6829$ & $1.6159$ & $0.5881$ & $0.0030$ & $2.1983$ & $-0.2145$ & $2.0058$ & $1.3676$ & $-0.6717$ \\ \hline
         & \multirow{2}{*}{T1}  &mean       & $0.9312$ & $4.8785$ & $20.2849$ & $587.5578$ & $4.8739$ & $17.8681$ & $10833.5256$ & $0.9763$ & $568.3542$ \\
         &                      &std        & $(0.0521)$ & $(0.5738)$ & $(0.0537)$ & $(72.4783)$ & $(0.3930)$ & $(5.5267)$ & $(3463.6525)$ & $(0.0422)$ & $(85.7772)$ \\ \cline{2-12}
MFEA     & \multirow{2}{*}{T2}  &mean       & $294.6075$ & $316.0845$ & $4450.4887$ & $116.8460$ & $8843.0128$ & $16.1079$ & $371.8389$ & $25.6697$ & $4609.5203$ \\
         &                      &std        & $(40.8627)$ & $(29.1948)$ & $(646.8284)$ & $(22.8572)$ & $(2198.8143)$ & $(5.1747)$ & $(49.9547)$ & $(3.4795)$ & $(495.9894)$ \\ \cline{2-12}
         & Score                &           & $0.4379$ & $0.1710$ & $-1.3312$ & $1.7700$ & $0.1219$ & $1.9218$ & $-0.0526$ & $0.4679$ & $0.0027$ \\ \hline
         & \multirow{2}{*}{T1}  &mean       & $0.9779$ & $4.8817$ & $20.2536$ & $439.1479$ & $5.2223$ & $8.2290$ & $19331.3190$ & $0.9241$ & $448.3269$ \\
         &                      &std        & $(0.0420)$ & $(0.3708)$ & $(0.0752)$ & $(74.1382)$ & $(0.4816)$ & $(6.1517)$ & $(8251.8875)$ & $(0.0865)$ & $(49.1185)$ \\ \cline{2-12}
MFEARR   & \multirow{2}{*}{T2}  &mean       & $328.0556$ & $341.4237$ & $4343.4306$ & $86.2004$ & $16927.1056$ & $15.5357$ & $419.0696$ & $36.7019$ & $4286.2454$ \\
         &                      &std        & $(30.5211)$ & $(44.2208)$ & $(586.3652)$ & $(27.3131)$ & $(6844.0965)$ & $(2.9138)$ & $(33.8599)$ & $(4.8207)$ & $(517.4152)$ \\ \cline{2-12}
         & Score                &           & $0.8810$ & $0.3928$ & $-1.4646$ & $0.2216$ & $1.0270$ & $0.5497$ & $1.0601$ & $1.3404$ & $-0.6591$ \\ \hline
         & \multirow{2}{*}{T1}  &mean       & $0.6213$ & $5.6270$ & $21.1512$ & $633.4187$ & $3.5734$ & $4.4146$ & $4250.2864$ & $1.0077$ & $1472.9144$ \\
         &                      &std        & $(0.1302)$ & $(0.8936)$ & $(0.1043)$ & $(126.1722)$ & $(0.3997)$ & $(0.8332)$ & $(1591.8700)$ & $(0.0352)$ & $(377.9349)$ \\ \cline{2-12}
LDA-MFEA & \multirow{2}{*}{T2}  &mean       & $252.7188$ & $307.0645$ & $8187.2562$ & $22.4887$ & $723.6025$ & $4.2403$ & $326.0426$ & $14.9481$ & $6773.0865$ \\
         &                      &std        & $(67.8589)$ & $(78.7763)$ & $(2356.7227)$ & $(9.0959)$ & $(211.2468)$ & $(1.2617)$ & $(69.9058)$ & $(1.8482)$ & $(882.5549)$ \\ \cline{2-12}
         & Score                &           & $-0.9739$ & $0.5781$ & $2.8382$ & $-0.1850$ & $-1.3942$ & $-1.6682$ & $-0.9915$ & $-0.4437$ & $4.7109$ \\ \hline
         & \multirow{2}{*}{T1}  &mean       & $0.9033$ & $4.8463$ & $20.2795$ & $577.3551$ & $5.0222$ & $20.1757$ & $11263.0957$ & $0.9909$ & $545.5573$ \\
         &                      &std        & $(0.0602)$ & $(0.4312)$ & $(0.0693)$ & $(86.9941)$ & $(0.4057)$ & $(0.0795)$ & $(3922.9374)$ & $(0.0296)$ & $(75.9845)$ \\ \cline{2-12}
MFEA-LBS & \multirow{2}{*}{T2}  &mean       & $304.3237$ & $319.2167$ & $4404.1627$ & $111.6610$ & $9756.5485$ & $19.3068$ & $393.8628$ & $25.1179$ & $4424.1310$ \\
         &                      &std        & $(42.9904)$ & $(39.1217)$ & $(620.6942)$ & $(22.3018)$ & $(3668.2769)$ & $(2.1333)$ & $(68.9549)$ & $(3.8481)$ & $(550.6847)$ \\ \cline{2-12}
         & Score                &           & $0.4254$ & $0.1773$ & $-1.3714$ & $1.5914$ & $0.2937$ & $2.7136$ & $0.1679$ & $0.4639$ & $-0.2499$ \\ \hline
         & \multirow{2}{*}{T1}  &mean       & $0.8927$ & $5.4012$ & $21.1841$ & $415.1442$ & $5.1704$ & $5.6786$ & $26599.1474$ & $0.9141$ & $450.3920$ \\
         &                      &std        & $(0.0730)$ & $(1.4196)$ & $(0.0455)$ & $(43.3671)$ & $(0.5054)$ & $(2.1553)$ & $(12441.1179)$ & $(0.0509)$ & $(51.0178)$ \\ \cline{2-12}
EBS      & \multirow{2}{*}{T2}  &mean       & $412.7935$ & $401.5351$ & $4121.1952$ & $89.3588$ & $26750.3202$ & $13.6764$ & $416.4035$ & $34.4474$ & $4245.4670$ \\
         &                      &std        & $(48.4015)$ & $(49.4289)$ & $(773.8833)$ & $(20.2276)$ & $(10428.6188)$ & $(2.9505)$ & $(43.3474)$ & $(7.9360)$ & $(569.8224)$ \\ \cline{2-12}
         & Score                &           & $1.3101$ & $1.2505$ & $0.3941$ & $0.1614$ & $1.8235$ & $-0.0717$ & $1.6515$ & $1.0940$ & $-0.6951$ \\ \hline
         & \multirow{2}{*}{T1}  &mean       & $0.9015$ & $4.8973$ & $\mathbf{20.2362}$ & $538.3215$ & $4.9770$ & $13.7719$ & $12605.2897$ & $0.9645$ & $579.1136$ \\
         &                      &std        & $(0.0679)$ & $(0.4284)$ & $(0.0647)$ & $(63.7993)$ & $(0.4259)$ & $(9.0053)$ & $(5139.6811)$ & $(0.0483)$ & $(106.9839)$ \\ \cline{2-12}
G-MFEA    & \multirow{2}{*}{T2}  &mean       & $298.4773$ & $325.5007$ & $\mathbf{3790.2887}$ & $101.2819$ & $9254.1989$ & $10.8155$ & $395.5002$ & $24.2534$ & $\mathbf{3972.5676}$ \\
         &                      &std       & $(32.0489)$ & $(37.5079)$ & $(565.2324)$ & $(24.6364)$ & $(3145.3736)$ & $(7.1813)$ & $(45.4696)$ & $(4.2472)$ & $(503.3188)$ \\ \cline{2-12}
         & Score                &           & $0.3695$ & $0.2648$ & $\mathbf{-1.8440}$ & $1.1298$ & $0.2224$ & $0.5746$ & $0.2950$ & $0.2950$ & $-0.6193$ \\ \hline
         & \multirow{2}{*}{T1}  &mean       & $0.7180$ & $3.9971$ & $21.2144$ & $398.5276$ & $4.0135$ & $3.8051$ & $4815.9985$ & $0.8285$ & $422.8324$ \\
         &                      &std        & $(0.0665)$ & $(0.4488)$ & $(0.0268)$ & $(49.3249)$ & $(0.2851)$ & $(0.3574)$ & $(2148.7461)$ & $(0.0882)$ & $(47.2119)$ \\ \cline{2-12}
EMEA    & \multirow{2}{*}{T2}  &mean       & $391.7383$ & $386.5877$ & $4530.9415$ & $56.2079$ & $4667.8277$ & $6.0926$ & $370.2845$ & $24.1708$ & $4164.9409$ \\
         &                      &std        & $(68.6963)$ & $(47.1808)$ & $(760.3681)$ & $(11.7526)$ & $(1854.2606)$ & $(2.5838)$ & $(54.6887)$ & $(4.8453)$ & $(608.4902)$ \\ \cline{2-12}
         & Score                &           & $0.5359$ & $0.2108$ & $0.7125$ & $-0.7085$ & $-0.7801$ & $-1.4692$ & $-0.5737$ & $-0.1614$ & $-0.8535$ \\ \hline
         & \multirow{2}{*}{T1}  &mean       & $\mathbf{0.0168}$ & $\mathbf{0.9795}$ & $21.1927$ & $\mathbf{48.5233}$ & $\mathbf{0.3267}$ & $\mathbf{2.0780}$ & $\mathbf{226.9160}$ & $\mathbf{0.0070}$ & $\mathbf{220.2267}$ \\
         &                      &std        & $(0.0174)$ & $(0.6684)$ & $(0.0383)$ & $(13.0611)$ & $(0.4127)$ & $(0.4938)$ & $(404.0593)$ & $(0.0058)$ & $(63.5012)$ \\ \cline{2-12}
MTGA     & \multirow{2}{*}{T2}  &mean       & $\mathbf{60.1727}$ & $\mathbf{50.1522}$ & $5844.5964$ & $\mathbf{0.0011}$ & $\mathbf{187.5526}$ & $\mathbf{1.8856}$ & $\mathbf{59.2972}$ & $\mathbf{7.7735}$ & $4602.3325$ \\
         &                      &std        & $(19.1548)$ & $(15.1310)$ & $(719.3296)$ & $(0.0016)$ & $(339.1746)$ & $(1.8059)$ & $(19.6508)$ & $(7.0682)$ & $(492.2891)$ \\ \cline{2-12}
         & Score                &         & $\mathbf{-4.6689}$ & $\mathbf{-4.6611}$ & $1.4783$ & $\mathbf{-3.9836}$ & $\mathbf{-3.5124}$ & $\mathbf{-2.3361}$ & $\mathbf{-3.5624}$ & $\mathbf{-4.4237}$ & $\mathbf{-0.9648}$ \\ \hline
\end{tabular} \label{tab:resultsO}
\end{table*}

Fig.~\ref{fig:O} and Table~\ref{tab:resultsO} also show that MTGA did not achieve the best performance on Benchmarks~3 and 9. Table~III in \cite{Da2017} gives the inter-task similarity of the two tasks in each benchmark. The two tasks in Benchmarks~3 and 9 have the lowest inter-task similarity (0.0002 and 0.0016, respectively) among the nine benchmarks. This fact may explain why MTGA did not perform well on them: though MTGA explicitly considers the bias between two tasks, it is still based on the assumption that the two tasks are similar. When the two tasks are almost completely different, MTGA has difficulty transferring useful information between them. This is a problem deserving future research.

\begin{figure*}[h]\centering
\includegraphics[width=\linewidth,clip]{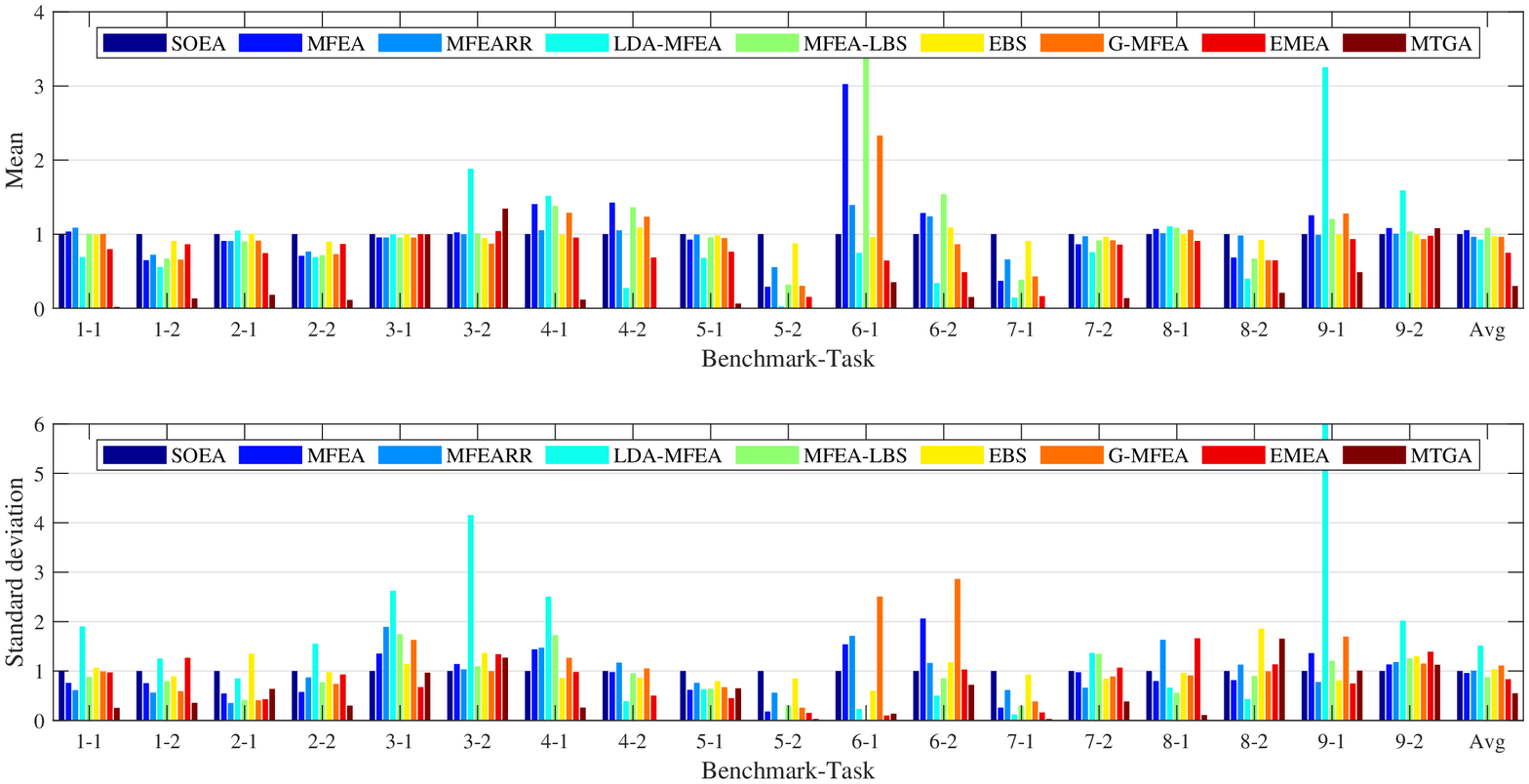}
\caption{Normalized mean and std of the errors of different algorithms (w.r.t. the SOEA) on the nine benchmarks, after 100,000 function evaluations.} \label{fig:std}
\end{figure*}

The normalized mean and std of different algorithms after 100,000 function evaluations, w.r.t. the SOEA, are shown in Fig.~\ref{fig:std}. Clearly, the MTGA achieved on average the smallest mean and std, suggesting that the MTGA consistently outperformed other approaches.

\begin{figure}[htpb]\centering
\includegraphics[width=\linewidth,clip]{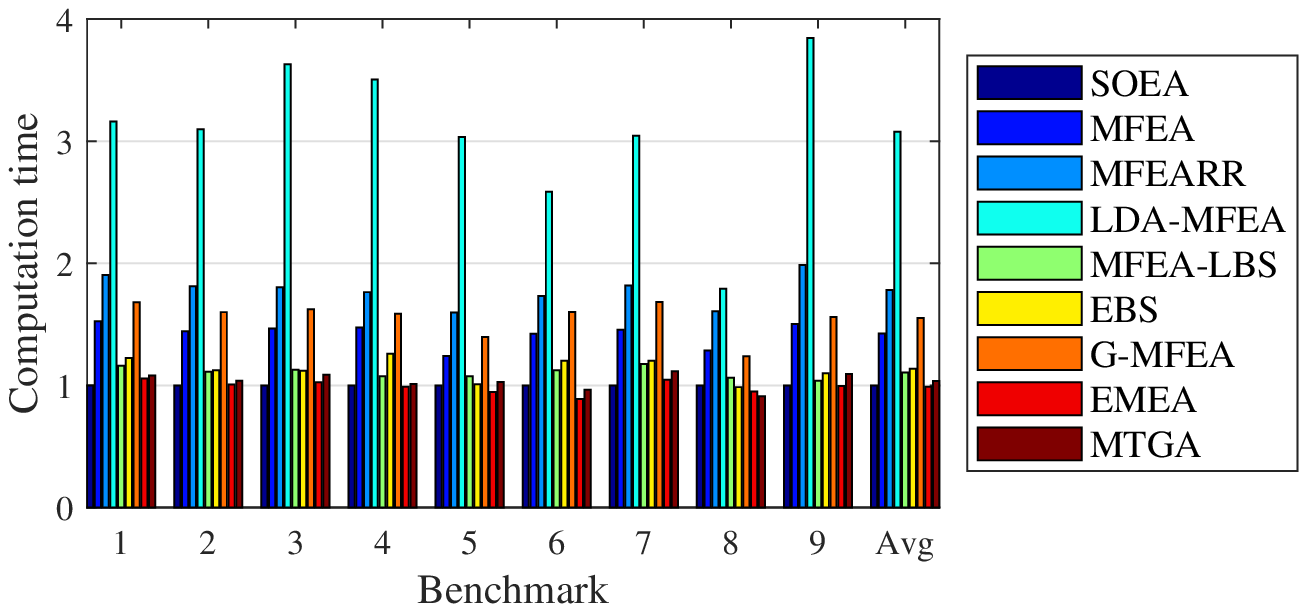}
\caption{Computation time of the nine algorithms.} \label{fig:OCompTime}
\end{figure}

It is also interesting to compare the computational cost of different algorithms. For each benchmark, we normalized the actual computation time (recorded in Matlab) of the eight multi-task algorithms w.r.t. the SOEA, and plot the results in Fig.~\ref{fig:OCompTime}. On average the SOEA, the EMEA and the MTGA had comparable computational cost (with values 1.0000, 0.9891, and 1.0368 respectively in the last group of Fig.~\ref{fig:OCompTime}), and they were the fastest among the nine. The MFEA-LBS and the EBS also had comparable computational cost, all of which were much smaller than the MFEA, the G-MFEA and the MFEARR. The computational cost of the LDA-MFEA was almost three times higher than the MTGA.

In summary, we can conclude that the proposed MTGA is effective and efficient.

\section{MTGA-Based Fuzzy System Optimization} \label{sect:FLCs}

This section applies the MTGA to fuzzy system optimization.

IT2 fuzzy systems have become very popular in the last two decades, because they have demonstrated outstanding performances in numerous applications \cite{drwuISA2006,drwuEAAI2006,Mendel2017,drwuBook2010}. Evolutionary algorithms, such as genetic algorithms, are frequently used to optimize IT2 fuzzy systems. There are generally two strategies for evolutionary IT2 fuzzy system optimization in the literature:
\begin{enumerate}
\item \emph{Partially dependent strategy}: An optimal T1 fuzzy system is designed first, and then its membership functions are blurred to obtain an IT2 fuzzy system.
\item \emph{Totally independent strategy}: An IT2 fuzzy system is optimized from scratch, without using a baseline T1 fuzzy system.
\end{enumerate}

Based on the MTGA, this section proposes a novel \emph{simultaneous optimization strategy} for fuzzy system design, which simultaneously optimize a T1 fuzzy system and an IT2 fuzzy system. This strategy has the following advantages:
\begin{enumerate}
\item Optimizing two fuzzy systems simultaneously by the MTGA may result in better fuzzy systems than optimizing each separately.
\item Obtaining a T1 fuzzy system and an IT2 fuzzy system simultaneously enables one to compare their performances and better determine which one to use in practice: if their performances are similar, then the T1 fuzzy system may be preferred for its simplicity; otherwise, the better-performing one (which is usually the IT2 fuzzy system) is chosen.
\end{enumerate}

All multi-tasking optimization algorithms compared in the previous section can be used in the \emph{simultaneous optimization strategy}. As an example, we compare MTGA with SOEA and MFEA in this section, on simultaneously optimizing T1 and IT2 FLCs for coupled-tank water level control \cite{drwuISA2006,drwuEAAI2006}. Their parameters were identical to those used in the previous section. Each algorithm was again run 20 times.

\subsection{The Coupled-Tank Water Level Control System}\label{sect:appendix}

\begin{figure}[!htbp]  \centering
  \includegraphics[width=\linewidth] {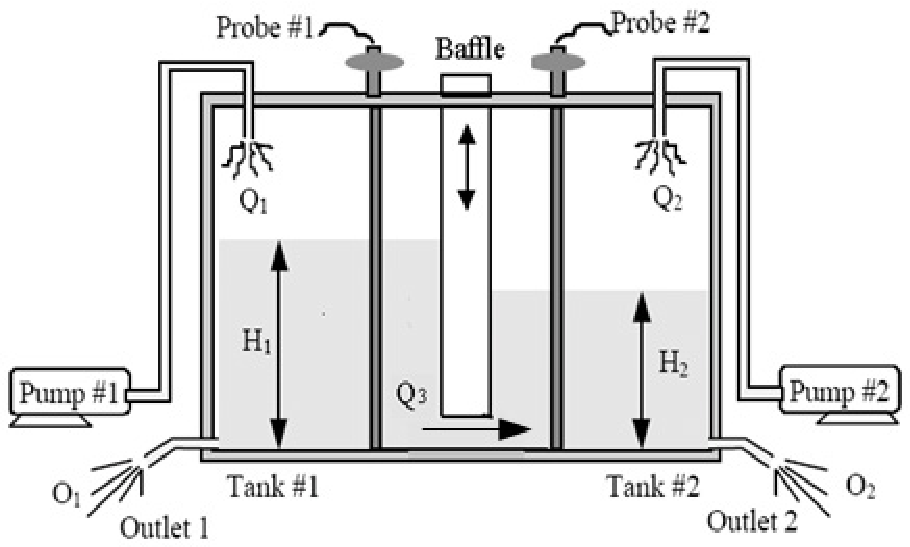}
\caption{The coupled-tank water-level control system.}
\label{fig:CTplant}
\end{figure}

T1 and IT2 FLCs were optimized to control the water level of the coupled-tank system shown in Fig.~\ref{fig:CTplant}. The plant has two small tower-type tanks mounted above a reservoir. Water can be pumped into the top of each tank by two independent pumps. The water levels are measured by two capacitive-type probe sensors. Each tank has an outlet, which allows water to flow out. Raising the baffle between the two tanks allows water to flow between them. The amount of water that returns to the reservoir is approximately proportional to the square root of the height of water in the tank, and hence the system is nonlinear.

The dynamics of the coupled-tank plant is described as:
\begin{align}
  A_1\frac{dH_1}{dt} &= Q_{1}-\alpha_1 \,\sqrt[]{H_1}-\alpha_3   \,\sqrt[]{H_1-H_2} \\
  A_2\frac{dH_2}{dt} &= Q_{2}-\alpha_2 \,\sqrt[]{H_2}+\alpha_3   \,\sqrt[]{H_1-H_2}   \label{eq:ct2}
\end{align}
where $A_1$ and $A_2$ are the cross-sectional areas of Tanks $\#1$ and $\#2$; $H_1$ and $H_2$ are the liquid levels in Tanks $\#1$ and $\#2$; $Q_1$ and $Q_2$ are the volumetric flow rates ($\mathbf{cm^3/s}$) of Pumps $\#1$ and $\#2$; $\alpha_1$, $\alpha_2$ and $\alpha_3$ are the proportionality constants corresponding to the $\sqrt{H_1}$, $\sqrt{H_2}$ and $\sqrt{H_1-H_2}$ terms, respectively.

In our experiment, Outlet~1 and Pump $\#2$ were shut off, and Pump $\#1$ was used to control the water level in Tank~$\#2$. In this case, $H_1 \geq H_2$ always held.

\subsection{Fitness Function}

Each chromosome in the population was evaluated on four different configurations of the couple-tank plant, shown in Table~\ref{tab:4plants}. This enables the optimized FLC to cope with a wide range of modeling uncertainties, and hence has a better chance to perform well on the actual plant. The same strategy was also used in our previous research \cite{drwuEAAI2006,drwuISA2006}.

\begin{table}[htpb]
    \caption{Four configurations of the water-level control plant for fitness evaluation.}   \label{tab:4plants}
    \setlength{\tabcolsep}{1.9mm}  \centering
  \begin{tabular}{lcccc}
  \toprule[1.5pt]
    & I & II & III & IV \\  \midrule[1pt]
    $A_1 = A_2$ (cm$^2$) & 36.52 & 36.52 & 36.52 & 36.52 \\
    $\alpha_1 = \alpha_2$ & 5.6186 & 5.6186 & 5.6186 & 5.6186 \\
    $\alpha_3$ & 10 & 10 & 10 & 8 \\
    Setpoint (cm) & 0 $\rightarrow$ 15 &  0 $\rightarrow$ 15 &
      0 $\rightarrow$ 22.5     $\rightarrow$ 7.5 & 0 $\rightarrow$ 15 \\
    Time delay (s) & 0 & 2 & 0 & 0 \\ \bottomrule[1.5pt]
  \end{tabular}
\end{table}

The fitness of each chromosome was the inverse of the sum of the integral of the time-weighted absolute errors (ITAEs) on the four plants \cite{drwuEAAI2006, drwuISA2006}:
\begin{align}
  F=\left[\sum_{p=1}^{4}w_p\left(\sum_{t=1}^{N_p}{t\cdot |e_p(t)|}\right)\right]^{-1},   \label{eq:ITAE2}
\end{align}
where $e_p(t)$ is the difference between the setpoint and the actual water height at the $t$th sampling of the $p$th plant, $w_p$ is the weight corresponding to the ITAE of the $p$th plant, and $N_p=200$ is the number of samples. Because the ITAE of the third plant is usually several times bigger than others, to balance the contributions from the four plants, $\alpha_3$ was defined as $1/3$ while the other three weights were unity.

\subsection{FLC Structure and Parameters}

\begin{figure*}[htpb]\centering
\includegraphics[width=.9\linewidth,clip]{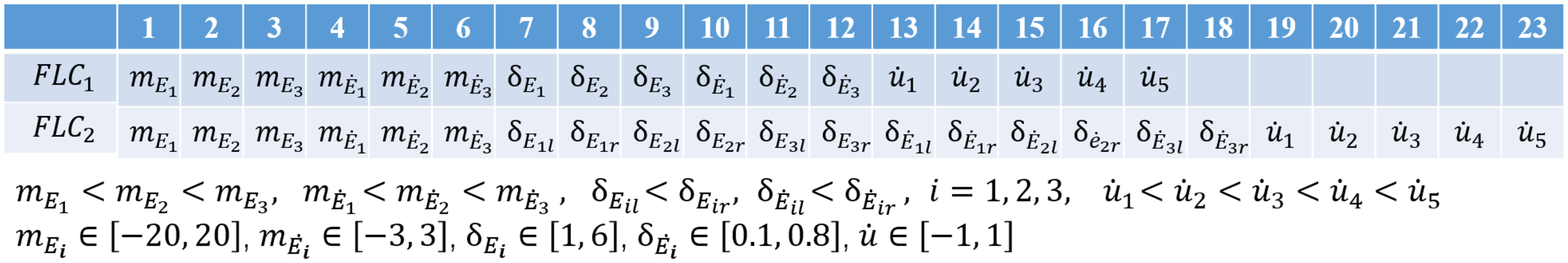}
\caption{Coding scheme of the T1 and IT2 FLCs.} \label{fig:gene}
\end{figure*}

Proportional-integral FLCs were used. The inputs were the feedback error $e$ and the change of error $\dot{e}$. The output was the change of control signal, $\dot{u}$. Three Gaussian membership functions were used in each input domain. The rulebase is shown in Table~\ref{tab:rules}, which included only five different consequents, $\dot{u}_i$ ($i = 1, 2, ..., 5$). Each T1 Gaussian membership function had two parameters: mean ($m$) and std ($\delta$). Each IT2 Gaussian membership function with a fixed mean and uncertain std had three parameters: mean ($m$), and the bounds of the std ($[\delta_l,\delta_r]$). So, the T1 FLC ($FLC_1$) had $2\times3\times2+5=17$ parameters, and the IT2 FLC ($FLC_2$) had $3\times3\times2+5=23$ parameters. The coding scheme is shown in Fig.~\ref{fig:gene}.

\begin{table}[htpb]
  \centering   \setlength{\tabcolsep}{5mm}
  \caption{The fuzzy rulebase used by both T1 and IT2 FLCs.}\label{tab:rules}
  \begin{tabular}{c|ccc}
  \toprule[1pt]
    $e\backslash \dot{e}$ &  $\dot{E}_1$ &  $\dot{E}_2$&     $\dot{E}_3$ \\ \midrule[.6pt]
    $E_1$ &  $\dot{u}_1$  &  $\dot{u}_2$ &  $\dot{u}_3$ \\
    $E_2$ &  $\dot{u}_2$ & $\dot{u}_3$ & $\dot{u}_4$ \\
    $E_3$ & $\dot{u}_3$ & $\dot{u}_4$ & $\dot{u}_5$ \\  \bottomrule[1pt]
  \end{tabular}
\end{table}

\begin{figure*}[htpb]\centering
\includegraphics[width=.7\linewidth,clip]{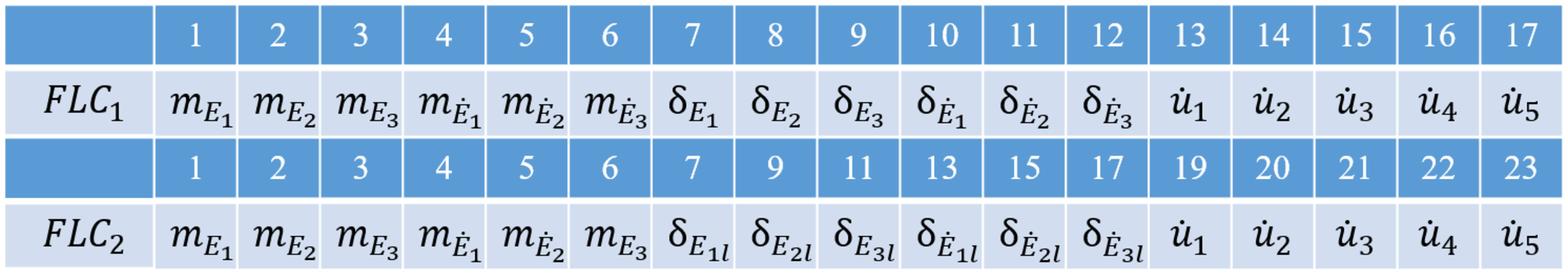}
\caption{Gene matching relationship in chromosome transfer between the T1 and IT2 FLCs.} \label{fig:match}
\end{figure*}

There were some natural constraints on the relative values of the parameters, as shown in Fig.~\ref{fig:gene}, e.g., $m_{E_1} < m_{E_2} < m_{E_3}$, where $m_{E_1}$, $m_{E_2}$ and $m_{E_3}$ are the mean of the Gaussian membership function of $E_1$, $E_2$ and $E_3$, respectively. So, we need to re-rank the genes to satisfy the constraints before evaluating the fitness of each chromosome.

In the previous section, MTGA used random matching in chromosome transfer. However, in this application, each parameter in the two FLCs has a specific physical meaning, which can be used to find the correspondence between parameters of the two FLCs. So, we used fixed matching in chromosome transfer, as shown in Fig.~\ref{fig:match}. The two genes in the same column were matched in chromosome transfer, and the genes in $FLC_2$ that do not appear in Fig.~\ref{fig:match} were not used in chromosome transfer. More specifically, $\delta_{E_{ir}}$ and $\delta_{\dot{E}_{ir}}$ ($i=1,2,3$) in the original $FLC_2$ chromosome were kept unchanged. However, because $\delta_{E_{il}}\le \delta_{E_{ir}}$ and $\delta_{\dot{E}_{il}}\le \delta_{\dot{E}_{ir}}$ should always be satisfied, when the transferred genes violated these constraints, the corresponding $\delta_{E_{il}}$ and $\delta_{E_{ir}}$ (or $\delta_{\dot{E}_{il}}$ and $\delta_{\dot{E}_{ir}}$) were switched.

It is also possible to transfer the average value of the upper and lower membership functions, instead of the upper or lower membership function only. This approach is similar to the \emph{partially dependent strategy} introduced at the beginning of this section, in which an optimal T1 fuzzy system is designed first, and then its membership functions are blurred to obtain an IT2 fuzzy system. A frequently used approach for blurring the std of the membership function is to change the std of a T1 membership function, e.g., $\delta$, to an interval $[\delta-\sigma,\delta+\sigma]$ centered at $\delta$. This chromosome transfer approach will be considered in our future research.

\subsection{Optimization Results}

\begin{figure}[htpb]\centering
\includegraphics[width=.9\linewidth]{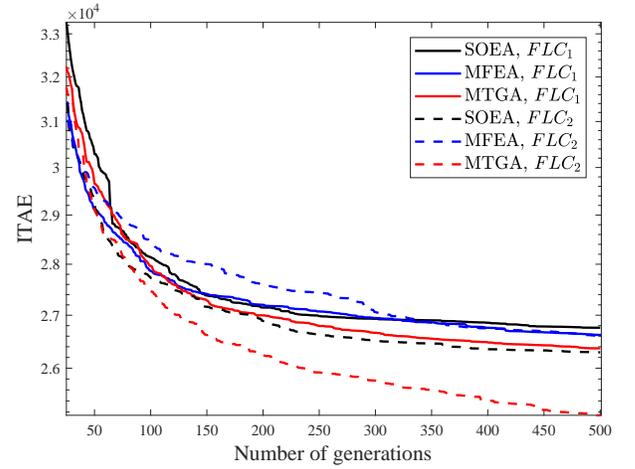}
\caption{ITAEs of the three optimization algorithms. To better visualize the differences, we only show the ITAEs after 25 generations.} \label{fig:FLCab}
\end{figure}

The optimization results of the three algorithms, averaged over 20 runs, are shown in Fig.~\ref{fig:FLCab}. We only plot the results after 25 generations so that they can be better distinguished.

Fig.~\ref{fig:FLCab} shows that:
\begin{enumerate}
\item When two SOEAs were used separately to optimize the T1 and IT2 FLCs, on average the IT2 FLC had smaller ITAE than the T1 FLC, which is consistent with many results in the literature \cite{drwuEAAI2006, drwuISA2006}, and also our expectation.
\item When the MFEA was used to optimize the T1 and IT2 FLCs simultaneously, $FLC_2$ may not outperform $FLC_1$, which is contradicting with our expectation, and suggests that MFEA may not be suitable for this real-world application.
\item When the MTGA was used to optimize the T1 and IT2 FLCs simultaneously, the IT2 FLC outperformed the T1 FLC, which coincided with the results in SOEA and also our expectation.
\item When the number of generation was large (e.g., larger than 125), the T1 (IT2) FLC obtained from the MTGA almost always gave a smaller ITAE than those obtained from the SOEA and the MFEA, suggesting that the MTGA is superior to the SOEA and the MFEA in real-world FLC optimization.
\end{enumerate}

\begin{figure}[htpb]\centering
\includegraphics[width=\linewidth,clip]{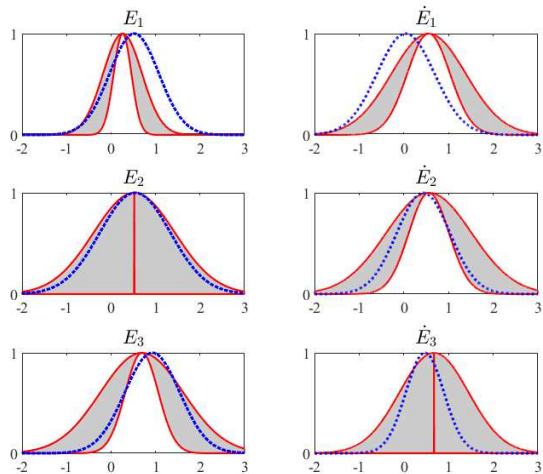}
\caption{Membership functions of the best T1 FLC (blue dotted curves) and IT2 FLC (red solid curves).} \label{fig:MFs}
\end{figure}

The membership functions of the best T1 and IT2 FLCs, obtained both from the MTGA, are shown in Fig.~\ref{fig:MFs}. The corresponding rule consequents are shown in Table~\ref{tab:bestRules}. It's interesting that $E_2$ and $\dot{E}_3$ of the IT2 FLC had almost completely filled-in footprint of uncertainties \cite{Mendel2017}, i.e., the lower membership functions were very narrow.

\begin{table}[htpb]   \centering   \setlength{\tabcolsep}{2mm}
  \caption{Rule consequents of the best T1 FLC and IT2 FLC (shown in the parentheses).} \label{tab:bestRules}
  \begin{tabular}{c|ccc}   \toprule[1pt]
    $e\backslash \dot{e}$ &  $\dot{E}_1$ &  $\dot{E}_2$&     $\dot{E}_3$ \\ \midrule[.6pt]
    $E_1$ &  $0.0338$ ($0.0636$) &  $0.1193$ ($0.3213$) &  $0.5345$ ($0.6013$)\\
    $E_2$ &  $0.1193$ ($0.3213$)& $0.5345$ ($0.6013$)& $0.7438$ ($0.7949$)\\
    $E_3$ & $0.5345$ ($0.6013$)& $0.7438$ ($0.7949$)& $0.9970$ ($0.8568$)\\  \bottomrule[1pt]
  \end{tabular}
\end{table}

In summary, we have demonstrated that when the MTGA is used, the newly proposed \emph{simultaneous optimization strategy} can be used to effectively optimize a T1 FLC and an IT2 FLC together.

\section{Conclusion} \label{sect:conclusions}

Evolutionary multi-tasking, or multi-factorial optimization, is an emerging subfield of multi-task optimization, which integrates evolutionary computation and multi-task learning to optimize multiple optimization tasks simultaneously. This paper has proposed a novel and easy-to-implement MTGA, which copes well with different tasks by estimating and using the bias among them. Comparative studies with eight existing approaches in the literature on nine benchmarks demonstrated that on average MTGA outperformed all of them, and had lower computational cost than six of them. Moreover, MTGA also demonstrated outstanding performance in simultaneous optimization of a T1 FLC and an IT2 FLC for coupled-tank water level control, which represents a brand-new \emph{simultaneous optimization strategy} for fuzzy system design. Although as an example we only demonstrated the effectiveness of the MTGA-based simultaneous optimization strategy in FLC optimization, it could also be used in other applications of fuzzy systems, e.g., classification and regression.


\end{document}